\documentclass[runningheads]{llncs}

\usepackage{amsmath}
\usepackage{amssymb}
\usepackage{amsfonts}
\usepackage{algorithmic}
\usepackage{listings}

\usepackage{graphicx}
\graphicspath{{imgs/}}
\DeclareGraphicsExtensions{.png,.pdf,.jpg}
\usepackage{caption}
\usepackage{subcaption}

\usepackage{textcomp}
\usepackage{xcolor}

\usepackage{booktabs}
\usepackage{multirow}
\usepackage[inline]{enumitem}

\usepackage{hyperref}
\usepackage{xurl}
\usepackage{cite}
\newcommand{\refappendix}[1]{\hyperref[#1]{Appendix~\ref*{#1}}}

\usepackage{notation}
\usepackage{stqlnotation}

\makeatletter
\renewcommand\paragraph{\@startsection{paragraph}{4}{\z@}%
  {-12\p@ \@plus -4\p@ \@minus -4\p@}%
  {-0.5em \@plus -0.22em \@minus -0.1em}%
  {\normalfont\normalsize\bfseries}}
\makeatother

\newlist{todolist}{itemize}{2}
\setlist[todolist]{label=$\square$,before*={\color{red}}}
\usepackage{pifont}

\begin{document}

% \begin{noindent}
\title{%
  PerceMon: Online Monitoring for Perception Systems%
  % \thanks{
  % This work was supported by the National Science Foundation under grant no. NSF-CNS-2038666 and Toyota Research Institute North America.}%
}
%
%\titlerunning{Abbreviated paper title}
%
\author{%
  Anand Balakrishnan\inst{1} \and
  Jyotirmoy Deshmukh\inst{1} \and
  Bardh Hoxha\inst{2} \and
  Tomoya Yamaguchi\inst{2} \and
  Georgios Fainekos\inst{3}}
\authorrunning{A.\ Balakrishnan et al.}
\institute{%
  University of Southern California\\
  \email{\{anandbal, jdeshmuk\}@usc.edu}%
  \and
  TRINA, Toyota Motor NA R\&D\\
  \email{\{bardh.hoxha, tomoya.yamaguchi\}@toyota.com}%
  \and
  Arizona State University\\
  \email{fainekos@asu.edu}
}
%\end{noindent}

\maketitle
\setcounter{footnote}{0}
\begin{abstract}

  Perception algorithms in autonomous vehicles are vital for the vehicle to understand
  the semantics of its surroundings, including detection and tracking of objects in the
  environment. The outputs of these algorithms are in turn used for decision-making in
  safety-critical scenarios like collision avoidance, and automated emergency braking.
  Thus, it is crucial to monitor such perception systems at runtime. However, due to the
  high-level, complex representations of the outputs of perception systems, it is a
  challenge to test and verify these systems, especially at runtime. In this paper, we
  present a runtime monitoring tool, PerceMon that can monitor arbitrary specifications
  in Timed Quality Temporal Logic (TQTL) and its extensions with spatial operators.
  % In this paper, we present an extension to Timed Quality Temporal Logic (TQTL) with
  % spatial set operations that can be used to express properties on high-level spatial
  % artifacts from perception systems, along with a runtime monitoring tool, PerceMon,
  % that can monitor such specifications at runtime.
  We integrate the tool with the CARLA autonomous vehicle simulation environment and the
  ROS middleware platform while monitoring properties on state-of-the-art object
  detection and tracking algorithms.

\end{abstract}

\section{Introduction}%
\label{sec:introduction}
%! TEX root = ../main.tex

In recent years, the popularity of autonomous vehicles has increased greatly. With this
popularity, there has also been increased attention drawn to the various fatalities
caused by the autonomous components on-board the vehicles, especially the perception
systems~\cite{templeton_tesla_2020,lee_report_2018}. Perception modules on these
vehicles use vision data from cameras to reason about the surrounding environment,
including detecting objects and interpreting traffic signs, and in-turn used by
controllers to perform safety-critical control decisions, including avoiding
pedestrians. Due to the nature of these systems, it has become important that these
systems be tested during design and monitored during deployment.

Signal temporal logic (STL)~\cite{maler_monitoring_2004} and Metric Temporal Logic
(MTL)~\cite{fainekos_robustness_2009} have been used extensively in verification,
testing, and monitoring of safety-critical systems. In these scenarios, typically there
is a model of the system that is generating trajectories under various actions. These
traces are the used to test if the system satisfies some specification. This is
referred to as \emph{offline monitoring}, and is the main setting for testing and
falsification of safety-critical systems. On the other hand, STL and MTL have been used
for \emph{online monitoring} where some safety property is checked for compliance at
runtime~\cite{nickovic_rtamt_2020,dokhanchi_online_2014}. These are used to express
rich specifications on low-level properties of signals outputted from systems.
%But, this expressivity does not extend to data from perception systems.

The output of a perception algorithm consists of a sequence of frames, where each frame
contains a variable number of objects over a fixed set of categories, in addition to
object attributes that can range over larger data domains (e.g.\ bounding box
coordinates, distances, confidence levels, etc.). STL and MTL can handle mixed-mode
signals and there have been attempts to extend them to incorporate spatial
data~\cite{bortolussi_specifying_2014,nenzi_qualitative_2015,haghighi_spatel_2015}.
However, these logics lack the ability to compare objects in different frames, or model
complex spatial relations between objects.

% When a perception algorithm outputs its detections, it does so in a mixed-mode setting:
% the trace contains discrete-valued signals (including object categories and IDs) as
% well as real-valued signals (namely, the bounding box coordinates and the confidence
% associated with a detection).
% %STL and MTL cannot handle such mixed-mode signals.
% While there have been attempts to extend STL and MTL to incorporate spatial
% data~\cite{bortolussi_specifying_2014,nenzi_qualitative_2015,haghighi_spatel_2015},
% they cannot express requirements over spatial objects present in signals from
% perception algorithms.

Timed Quality Temporal Logic (TQTL)~\cite{dokhanchi_evaluating_2018}, and
Spatio-temporal Quality Logic (STQL)~\cite{hekmatnejad_formalizing_2021} are extensions
to MTL that incorporate the semantics for reasoning about data from perception systems
specifically. In STQL, which is in itself an extension of TQTL, the syntax defines
operators to reason about discrete IDs and classes of objects, along with set
operations on the spatial artifacts, like bounding boxes, outputted by perception
systems.

In this project, we contribute the following:
\begin{enumerate}

  \item We show how TQTL~\cite{dokhanchi_evaluating_2018} and
        STQL~\cite{hekmatnejad_formalizing_2021} can be used to express correctness properties
        for perception algorithms.

        % The use of Spatio-Temporal Quality Logic (STQL)~\cite{hekmatnejad_formalizing_2021}, a
        % spatial extension to TQTL~\cite{dokhanchi_evaluating_2018}: we formulate few properties
        % that aim to catch some common issues with perception algorithms.

  \item An online monitoring tool,
        \emph{PerceMon}\footnote{\url{https://github.com/CPS-VIDA/PerceMon.git}}, that
        efficiently monitors \stql{} specifications. We integrate this tool with the CARLA
        simulation environment~\cite{dosovitskiy_carla_2017} and the Robot Operating System
        (ROS)~\cite{quigley_ros_2009}.

\end{enumerate}

\begin{figure}[htpb]
  \centering
  \includegraphics[width=\textwidth]{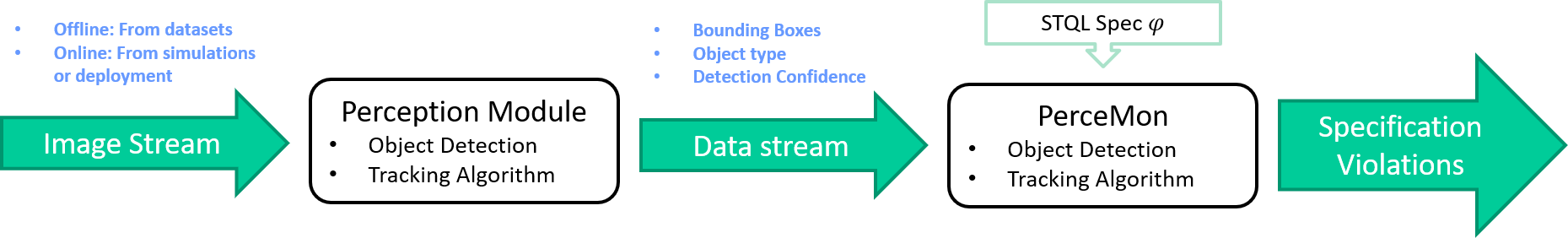}
  \caption{The PerceMon online monitoring pipeline.}%
  \label{fig:percemon-overview}
\end{figure}

\paragraph{Related work}%
S-TaLiRo~\cite{annpureddy_staliro_2011,fainekos_robustness_2019},
VerifAI~\cite{dreossi_verifai_2019} and Breach~\cite{donze_automotive_2015} are some
examples of tools used for offline monitoring of MTL and STL specifications. The
presented tool, \emph{PerceMon}, models its architecture similar to the
RTAMT~\cite{nickovic_rtamt_2020} online monitoring tool for STL specifications: the
core tool is written in C++ with an interface for use in different,
application-specific platforms.

\section{Spatio-temporal Quality Logic}%
\label{sec:stql}

Spatio-temporal quality logic (STQL)~\cite{hekmatnejad_formalizing_2021} is an
extension of Timed Quality Temporal Logic (TQTL)~\cite{dokhanchi_evaluating_2018} that
incorporates reasoning about high-level topological structures present in perception
data, like bounding boxes, and set operations over these structures.

% Need for spatio-temporal logics
STL has been used extensively in testing and monitoring of control systems mainly due
to the ability to express rich specifications on low-level, real-valued signals
generated from these systems. To make the logic more high-level, spatial extensions
have been proposed that are able to reason about spatial relations between
signals~\cite{bortolussi_specifying_2014,nenzi_qualitative_2015,gabelaia_computational_2003,haghighi_spatel_2015}.
A key feature of data streams generated by perception algorithms is that they contain
\emph{frames} of  spatial objects consisting of both, real-values and discrete-valued
quantities: the discrete-valued signals are the IDs of the objects and their associated
categories; while real-valued signals include bounding boxes describing the objects and
confidence associated with their identities. While STL and MTL can be used to reason
about properties of a fixed number of such objects in each frame by creating signal
variables to encode each of these properties, it is not possible to design monitors
that handle arbitrarily many objects per frame.
%STL and it's extensions are not able to handle such mixed-mode signals.

TQTL~\cite{dokhanchi_evaluating_2018} is a logic that is specifically catered for
spatial data from perception algorithms. Using Timed Propositional Temporal
Logic~\cite{bouyer_expressiveness_2005} as a basis, TQTL allows one to \emph{pin} or
\emph{freeze} the signal at a certain time point and use clock variables associated
with the freeze operator to define time constraints. Moreover, TQTL introduces a
quantifier over objects in a frame and the ability to refer to properties intrinsic to
the object: tracking IDs, classes or categories, and detection confidence.
STQL~\cite{hekmatnejad_formalizing_2021} further extends the logic to reason about the
bounding boxes associated with these objects, along with predicate functions for these
spatial sets, by incorporating topological semantics from the \(S4_u\) spatio-temporal
logic~\cite{gabelaia_computational_2003}.

\begin{definition}[STQL
    Syntax~\cite{hekmatnejad_formalizing_2021}]%
  \label{def:stql-syntax} Let \(V_t\) be a set of time variables, \(V_f\) be a set of
  frame variables, and \(V_o\) be a set of object ID variables. Then the syntax for STQL
  is recursively defined by the following grammar:
  \begin{align*}
    % Top-level expression
    \varphi ::=\quad%
                    & \exists\{ {id}_1, {id}_2, \ldots\}@\varphi \mid
    \{x,f\}.\varphi                                                         \\
                    & \mid \top \mid \neg\varphi \mid \varphi \lor \varphi
    \mid \Next\varphi \mid \Prev\varphi \mid \varphi \Unt \varphi \mid \varphi
    \Since \varphi                                                          \\
                    & \mid \CTIME - x \sim t \mid \CFRAME - f \sim n
    \\
                    & \mid \idcls({id}_i) = c \mid \idcls({id}_i) =
    \idcls({id}_i)
    \mid \idprob({id}_i) \geq r \mid \idprob({id}_i) \geq r \times
    \idprob({id}_j)                                                         \\
                    & \mid \{{id}_i = {id}_j\} \mid \{{id}_i \not= {id}_j\}
    \mid \SpExists \Omega \mid \Pi                                          \\
    \Omega ::=\quad & \varnothing \mid \Universal \mid \BBox({id}_1)
    \mid \Comp\Omega \mid \Omega \sqcup \Omega
    \\
    \Pi ::=\quad    & \Area(\Omega) \geq r | \Area(\Omega) \geq r \times
    \Area(\Omega)                                                           \\
                    & \mid \ED({id}_i, \CRT, {id}_j, \CRT) \geq r
    \mid \Theta \geq r \mid \Theta \geq r \times \Theta
    \\
    % Distance operations
    \Theta ::=\quad & \Lat({id}_i, \CRT) \mid \Lon({id}_i, \CRT)
    \\
    % Reference points
    \CRT ::=\quad   & \LM \mid \RM \mid \TM \mid \BM \mid \CT
  \end{align*}

  Here, \({id}_i \in V_o\) (for all indices \(i\)), \(x \in V_t\), and \(f \in V_f\). In
  the above grammar \(r\) is a real-valued constant that allows for the comparison of
  ratios of object properties.
\end{definition}

In the above grammar, \(\neg\varphi\) and \(\varphi \lor \varphi\) are, respectively,
the negation and disjunction operators from propositional logic while \(\Next\varphi\),
\(\Prev\varphi \), \( \varphi \Unt \varphi \), and \( \varphi \Since \varphi\) are the
temporal operators \emph{next}, \emph{previous}, \emph{until}, and \emph{since}
respectively. The above grammar can be further used to derive the other propositional
operators, like conjunction (\(\varphi \land \varphi\)), along with temporal operators
like \emph{always} (\(\Alw \varphi\)) and \emph{eventually} (\(\Ev \varphi\)), and
their past-time equivalents \emph{holds} (\(\AlwP \varphi\)) and \emph{once} (\(\EvP
\varphi\)). In addition to that, STQL extends these by introducing freeze quantifiers
over clock variables and object variables. \(\Set{x, f}.\varphi\) freezes the time and
frame that the formula \(\varphi\) is evaluated, and assigns them to the clock
variables \(x\) and \(f\), where \(x\) refers to pinned time variables and \(f\) refers
to pinned frame variables. The constants, \(\CTIME, \CFRAME\) refer to the value of the
time and frame number where the current formula is being evaluated. This allows for the
expression \(x - \CTIME\) and \(f - \CFRAME\) to measure the duration and the number of
frames elapsed, respectively, since the clock variables \(x\) and \(f\) were pinned.
The expression \(\exists\Set{{id}_1}@\varphi\) searches over each object in a frame in
the incoming data stream --- assigning each object to the object variable \({id}_1\)
--- if there exists an object that satisfies \(\varphi\). The functions \(\idcls(id)\)
and \(\idprob(id)\) refer to the \emph{class} and \emph{confidence} the detected object
associated with the ID variable. In addition to these TQTL operations, bounding boxes
around objects can be extracted using the expression \(\BBox(id)\) and set topological
operations can be defined over them. The \emph{spatial exists} operator
\(\SpExists\Omega\) checks if the spatial expression \(\Omega\) results in a non-empty
space or not. Quantitative operations like \(\Area(\cdot)\) measure the area of spatial
sets; \(\ED\) computes the Euclidean distances between references points (\(\CRT\)) of
bounding boxes; and \(\Lat\) and \(\Lon\) measure the latitudinal and longitudinal
offset of bounding boxes respectively. Here, \(\CRT\) refers to the reference points
--- left, right, top, and bottom margins, and the centroid --- for bounding boxes. Due
to lack of space, we defer defining the formal semantics of STQL
to~\refappendix{sec:semantics_for_stql} and also refer the readers
to~\cite{hekmatnejad_formalizing_2021} for more extensive details.

% Similar to~\cite{dokhanchi_online_2014}, we define a syntactical notion of history and horizon that quantifies the maximum num
% \begin{definition}[Syntactically Bounded History and Horizon]%
%   \label{def:horizon}
%
% \end{definition}

% \section{Past-time formalization for STQL}%
% \label{sec:past-time}
% \input{sections/past-time.tex}

\section{PerceMon: An Online Monitoring Tool}%
\label{sec:percemon}
%! TEX root = ../main.tex

\begin{figure}[hbtp]
  \centering
  \begin{subfigure}[t]{0.49\textwidth}
    \begin{center}
      \includegraphics[width=\textwidth]{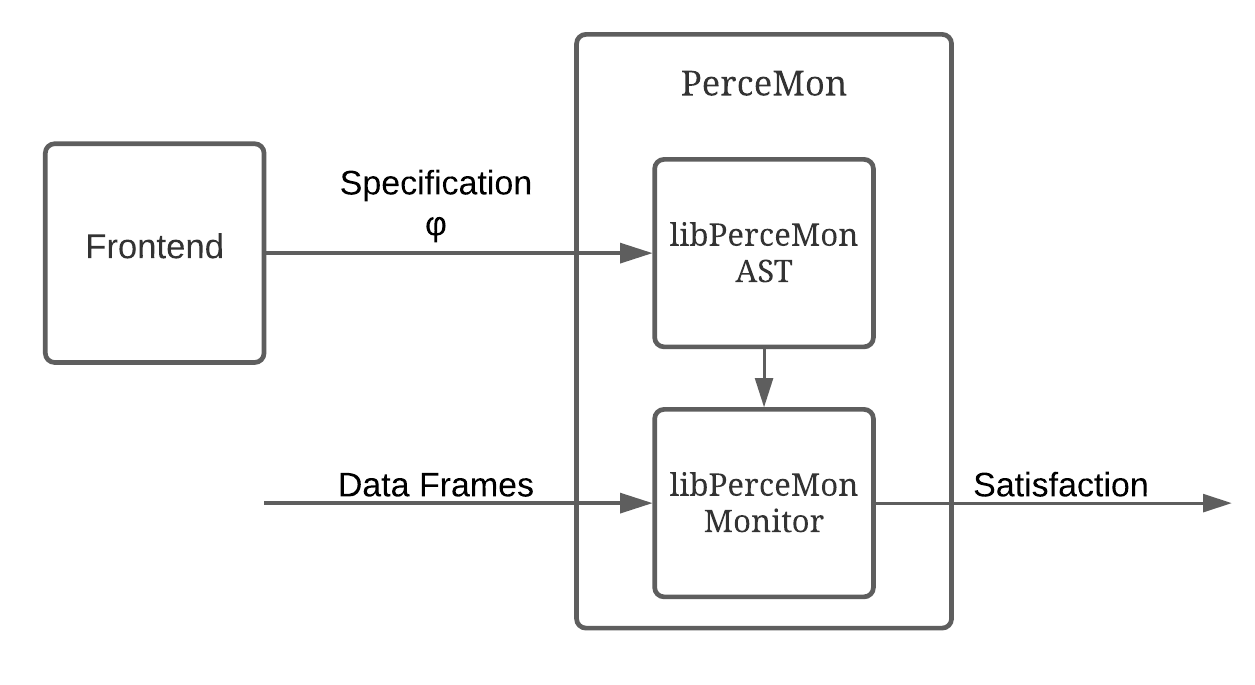}
    \end{center}
    \caption{General architecture for PerceMon. The \emph{frontend} component is a
      generic wrapper around \texttt{libPerceMon}, the C++ library that provides the online
      monitoring functionality, for example, a wrapper for ROS, a parser from some
      specification language, or a Python library.}%
    \label{fig:percemon-architecture}
  \end{subfigure}
  \hfill
  \begin{subfigure}[t]{0.49\textwidth}
    \begin{center}
      \includegraphics[width=\textwidth]{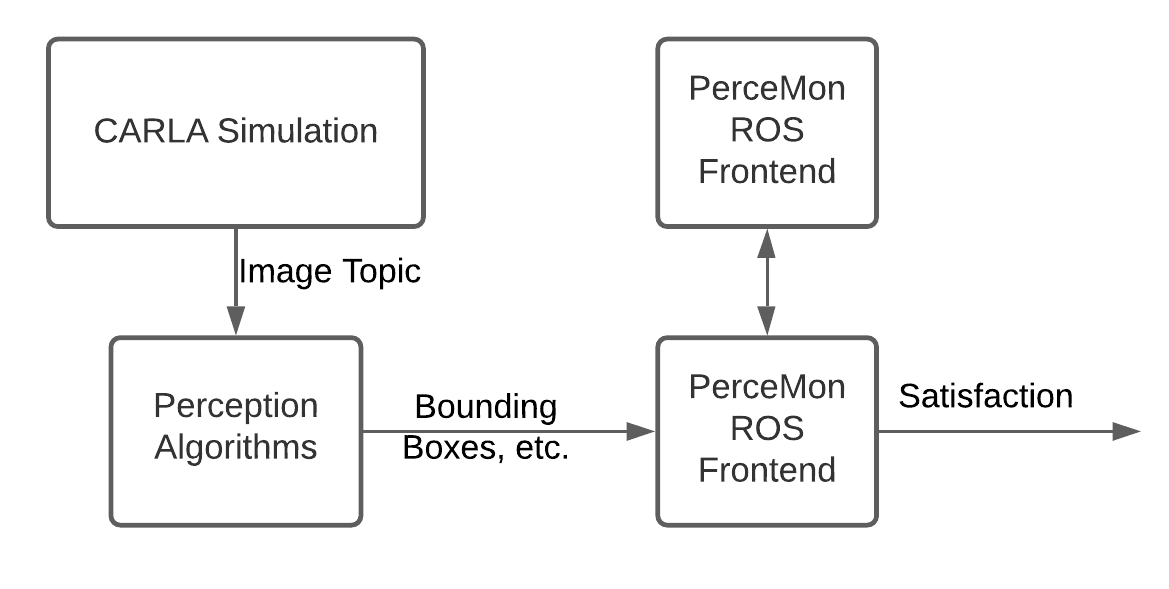}%
    \end{center}
    \caption{Architecture of the integration of PerceMon with the CARLA
      autonomous vehicle simulator and ROS middleware
      platform.}%
    \label{fig:percemon-carla-ros}%
  \end{subfigure}
\end{figure}

\emph{PerceMon} is an online monitoring tool for STQL specifications. It computes the
quality of a formula \(\varphi\) at the current evaluation frame, if \(\varphi\) can be
evaluated with some finite number of frames in the past (\emph{history}) and delayed
frames from the future (\emph{horizon}).

The core of the tool consists of a C++ library, \texttt{libPerceMon}, which provides an
interface to define an STQL abstract syntax tree efficiently, along with a general
online monitor interface. The \emph{PerceMon} tool works by initializing a monitor with
a given \stql{} specification and can receive data in a frame-by-frame manner. It
stores the frames in a first-in-first-out (FIFO) buffer with maximum size defined by
the horizon and history requirement of the specification. This enables fast and
efficient computation of the quality of the formula for the bounded horizon. An
overview of the architecture can be seen in~\autoref{fig:percemon-architecture}.

The library, \texttt{libPerceMon}, designed with the intention to be used with wrappers
that convert application-specific data to data structures supported by the library
(signified by the ``Frontend'' block in the architecture presented
in~\autoref{fig:percemon-architecture}). In the subsequent section, we show an example
of how such an integration can be performed by interfacing \texttt{libPerceMon} with
the CARLA autonomous vehicle simulator~\cite{dosovitskiy_carla_2017} via the ROS
middleware platform~\cite{quigley_ros_2009}.

\subsection{Integration with CARLA and ROS}%
\label{sub:integration_with_carla_and_ros}

% \begin{figure}[htpb]
%   \centering
%   \includegraphics[width=0.65\linewidth]{percemon-carla-ros}
%   \caption{Architecture of the integration of PerceMon with the CARLA autonomous
%     vehicle simulator and ROS middleware platform.}%
%   \label{fig:percemon-carla-ros}
% \end{figure}

In this section, we present an integration of the \emph{PerceMon} tool with the CARLA
autonomous vehicle simulator~\cite{dosovitskiy_carla_2017} using the ROS middleware
platform~\cite{quigley_ros_2009}. This follows the example
of~\cite{dreossi_verifai_2019} and~\cite{zapridou_runtime_2020} which interface with
CARLA, and~\cite{nickovic_rtamt_2020}, where the tool interfaces with the ROS
middleware platform for use in online monitoring applications.

The CARLA simulator is an autonomous vehicle simulation environment that uses
high-quality graphics engines to render photo-realistic scenes for testing such
vehicles. Pairing this with ROS allows us to abstract the data generated by the
simulator, the PerceMon monitor, and various perception modules as streams of data or
\emph{topics} in a publisher-subscriber network model. Here, a \emph{publisher}
broadcasts data in a known binary format at an endpoint (called a \emph{topic}) without
knowing who listens to the data. Meanwhile, a \emph{subscriber} registers to a specific
topic and listens to the data published on that endpoint.

In our framework, we use the ROS wrapper for
CARLA\footnote{\url{https://github.com/carla-simulator/ros-bridge/}} to publish all the
information from the simulator, including data from the cameras on the autonomous
vehicle. The image data is used by perception modules --- like the YOLO object
detector~\cite{redmon_you_2016} and the DeepSORT object
tracker~\cite{wojke_simple_2017} --- to publish processed data. The information
published by these perception modules can in-turn be used by other perception modules
(like using detected objects to track them), controllers (that may try to avoid
collisions), and by PerceMon online monitors. The architectural overview can be seen
in~\autoref{fig:percemon-carla-ros}.

The use of ROS allows us to reason about data streams independent of the programming
languages that the perception modules are implemented in. For example, the main
implementation of the YOLO object detector is written in C/C++ using a custom deep
neural network framework called \emph{Darknet}~\cite{darknet13}, while the DeepSORT
object tracker is implemented in Python. The custom detection formats from each of
these algorithms can be converted into standard messages that are published on
predefined topics, which are then subscribed to from PerceMon. Moreover, this also
paves the way to migrate and apply PerceMon to any other applications that use ROS for
perception-based control, for example, in the software stack deployed on real-world
autonomous vehicles~\cite{kato_autoware_2018}.

\section{Experiments}%
\label{sec:experiments}
%! TEX root = ../main.tex

\begin{figure}[htpb]
  \centering
  \begin{subfigure}[t]{0.49\textwidth}
    \centering
    \includegraphics[width=\textwidth]{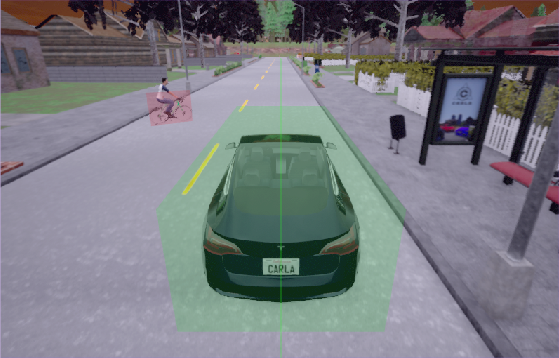}
    \caption{In this scenario, the configuration is such that the sun has set. In a
      poorly lit road, a cyclist tries to cross the road.}%
    \label{fig:uber-scenario}
  \end{subfigure}
  \hfill
  \begin{subfigure}[t]{0.49\textwidth}
    \centering \includegraphics[width=\textwidth]{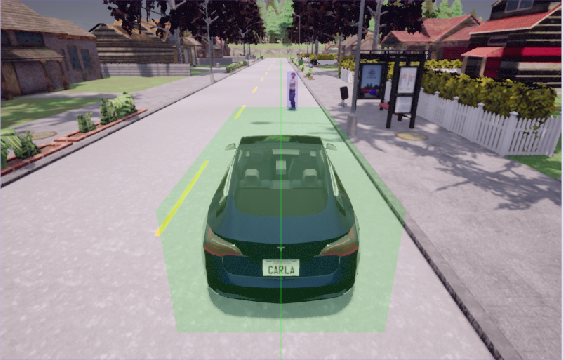} \caption{Here, a
      partially occluded pedestrian decides to suddenly cross the road as the vehicle cruises
      down the street.}%
    \label{fig:pedestrian-crossing-scenario}
  \end{subfigure}
  \caption{The presented
    scenarios simulated in CARLA aim to demonstrate some common failures associated with
    deep neural network-based perception modules. These include situations where partially
    occluded objects are not detected or tracked properly, and situations where different
    lighting conditions cause mislabeling of detected objects. In both the above scenarios,
    we also add some passive vehicles to increase the number of objects detected in any
    frame. This allows us to compute the time it takes to compute the satisfaction values
    from the monitor as the number of objects that need to be checked increases.}
  \label{fig:carla-experiments}%
\end{figure}

In this section, we present a set of experiments using the integration of
\emph{PerceMon} with the CARLA autonomous car simulator~\cite{dosovitskiy_carla_2017}
presented in Section~\ref{sub:integration_with_carla_and_ros}. We build on the
ROS-based architecture described in the previous section, and monitor the following
perception algorithms:
\begin{itemize}

  \item \emph{Object Detection}: The YOLO object
        detector~\cite{redmon_you_2016,redmon_yolov3_2018} is a deep convolutional neural
        networks (CNN) based model that takes as input raw images from the camera and outputs a
        list of bounding boxes for each object in the image.

  \item \emph{Object Tracking}: The SORT object tracker~\cite{wojke_simple_2017} takes
        the set of detections from the object detector and associates an ID with each of them.
        It then tries to track each annotated object across frames using Kalman filters and
        cosine association metrics.

\end{itemize}

We use the OpenSCENARIO specification format~\cite{openscenario_specification} to
define scenarios in the CARLA simulation that mimic some real-world, accident-prone
scenarios, where there have been several instances where deep neural network based
perception algorithms fail at detecting or tracking pedestrians, cyclists, and other
vehicles. To detect some common failure cases, we initialize the \emph{PerceMon}
monitors with the following specifications:

\paragraph{Consistent Detections $\varphi_1$}: For all objects in the current frame
that have high confidence, if the object is far away from the margins, then the object
must have existed in the previous frame too with sufficiently high confidence.
%\begin{noindent}
\begin{align}
  \begin{split}
    \varphi_1                  & := \forall\Set{id_1} @ \Set{f}.\left( \left(\varphi_{\text{high prob}} \land \varphi_{\text{margins}}\right) \Rightarrow \Prev\varphi_{\text{exists}}\right) \\
    \varphi_{\text{high prob}} & := \idprob(id_1) > 0.8 \\
    \varphi_{\text{margins}}   & := \Lon(id_1, \TM) > c_1 \land \Lon(id_1, \BM) < c_2 \\
                               & \quad\land \Lat(id_1, \LM) > c_3 \land \Lat(id_1, \RM) < c_4 \\
    \varphi_{\text{exists}}    & := \exists\{id_2\}.\left(\{id_1 = id_2\} \land \idprob(id_2) > 0.7\right)
  \end{split}
\end{align}
%\end{noindent}
Object detection algorithms are known to frequently miss detecting objects in
consecutive frames or detect them with low confidence after detecting them with high
confidence in previous frames. This can cause issues with algorithms that rely on
consistent detections, e.g., for obstacle tracking and avoidance. The above formula
checks this for objects that we consider ``relevant'' (using
\(\varphi_\text{margins}\)), i.e., the object is not too far away from the edges of the
image. This allows us to filter false alarms from objects that naturally leave the
field of view of the camera.

\paragraph{Smooth Object Trajectories $\varphi_2$}: For every object in the current
frame, its bounding box and the corresponding bounding box in the previous frame must
overlap more than 30\%.
\begin{align}
  \begin{split}
    \varphi_2              & :=
    \forall\Set{id_1} @ \Set{f_1}. \left( \Prev \left( \exists\Set{id_2} @ \Set{f_2} .
    \left( \Set{id_1 = id_2} \Rightarrow \varphi_\text{overlap} \right) \right) \right) \\
    \varphi_\text{overlap} & := \frac{\Area(\BBox(id_1) \sqcap
      \BBox(id_2))}{\Area(\BBox(id_1))} \geq 0.3
  \end{split}
\end{align}
In consecutive
frames, if detected bounding boxes are sufficiently far apart, it is possible for
tracking algorithms that rely on the detections to produce incorrect object
associations, leading to poor information for decision-making.

We monitor the above properties for scenarios described
in~\autoref{fig:carla-experiments}, and check for the time it takes to compute the
satisfaction values of the above properties. As each scenario consists of some passive
or non-adversarial vehicles, the number of objects detected by the object detector
increases. Thus, since the runtime for the STQL monitor is exponential in the number of
object IDs referenced in the existential quantifiers, this allows us to empirically
measure the amount of time it takes to compute the satisfaction value in the monitor.
The number of simulated non-adversarial objects are ranged from 1 to 10, and the time
taken to compute the satisfaction value for each new frame is recorded. We present the
results in~\autoref{tab:results}, and refer the readers
to~\cite{hekmatnejad_formalizing_2021} for theoretical results on monitoring complexity
for STQL specifications.

\begin{table}[htpb]
  \centering
  \caption{Compute time for different properties, with increasing number of
    objects.} \label{tab:results}

  %\begin{noindent}
  \begin{tabular*}{\textwidth}{@{\extracolsep{\fill}}ccc}
    \toprule Average Number of Objects & \multicolumn{2}{c}{Average Compute Time (s)}                 \\
    \cmidrule{2-3}              & \(\varphi_1\)                             & \(\varphi_2\) \\
    \midrule 2                  & \(7.0 \times 10^{-6}\)                    & \(7.3 \times 10^{-6}\) \\
    5                           & \(1.4 \times 10^{-5}\)                    & \(2.3 \times 10^{-5}\) \\
    10                          & \(5.4 \times 10^{-4}\)                    & \(6.3 \times 10^{-4}\) \\\bottomrule
  \end{tabular*}
  %\end{noindent}
\end{table}

\section{Conclusion}%
\label{sec:conclusion}

In this paper, we presented \emph{PerceMon}, an online monitoring library and tool for
generating monitors for specifications given in Spatio-temporal Quality Logic (STQL).
We also present a set of experiments that make use of \emph{PerceMon}'s integration
with the CARLA autonomous car simulator and the ROS middleware platform.

In future iterations of the tool, we hope to incorporate a more expressive version of
the specification grammar that can reason about arbitrary spatial constructs, including
oriented polygons and segmentation regions, and incorporate ways to formally reason
about system-level properties (like system warnings and control inputs).

\section*{Acknowledgment}

This work was partially supported by the National Science Foundation under grant no.
NSF-CNS-2038666 and the tool was developed with support from Toyota Research Institute
North America.

% \nocite{*}
\bibliographystyle{splncs04}
\bibliography{bib}

\newpage
\appendix
%! TEX root = ../main.tex

\section{Semantics for STQL}%
\label{sec:semantics_for_stql}

Consider a data stream \(\xi\) consisting of \emph{frames} containing objects and
annotated with a time stamp. Let \(i \in \Naturals\) be the current frame of
evaluation, and let \(\xi_i\) denote the \(i^{th}\) frame. We let \(\epsilon: V_t \cup
V_f \to \Naturals \cup \{\NaN\} \) denote a mapping from a pinned time or frame
variable to a frame index (if it exists), and let \(\zeta: V_o \to \Naturals\) be a
mapping from an object variable to an actual object ID that was assigned by a
quantifier. Finally, we let \(\Oc(\xi_i)\) denote the set of object IDs available in
the frame \(i\), and let \(t(\xi_i)\) output the timestamp of the given frame.

Let \(\Quality{\varphi}\) be the quality of the \stql{} formula, \(\varphi\), at the
current frame \(i\), which can be recursively defined as follows:

\begin{itemize}

  \item For the propositional and temporal operations, the semantics simply follows the
        Boolean semantics for LTL or MTL, i.e.,
        \begin{align*}
          \Quality{\top}(\xi, i, \epsilon,
          \zeta)                                                        & = \top
          \\
          \Quality{\neg\varphi}(\xi, i, \epsilon, \zeta)                & = \neg\Quality{\varphi}(\xi,
          i, \epsilon, \zeta)                                                                                                                          \\
          \Quality{\varphi_1 \lor \varphi_2}(\xi, i, \epsilon,
          \zeta)                                                        & = \Quality{\varphi_i}(\xi, i, \epsilon, \zeta) \lor \Quality{\varphi_2}(\xi,
          i, \epsilon, \zeta)                                                                                                                          \\
          \Quality{\Next\varphi}(\xi, i, \epsilon, \zeta)               & =
          \Quality{\varphi}(\xi, i + 1, \epsilon, \zeta)                                                                                               \\
          \Quality{\Prev\varphi}(\xi, i,
          \epsilon, \zeta)                                              & = \Quality{\varphi}(\xi, i - 1, \epsilon, \zeta)                             \\
          \Quality{\varphi_1 \Until \varphi_2}(\xi, i, \epsilon, \zeta) & = \bigvee_{i \leq
            j}\left( \Quality{\varphi_2}(\xi, j, \epsilon, \zeta) \land \bigwedge_{i \leq k \leq j}
          \Quality{\varphi_1}(\xi, k, \epsilon, \zeta) \right)
          \\
          \Quality{\varphi_1 \Since \varphi_2}(\xi, i,
          \epsilon, \zeta)                                              & = \bigvee_{j \leq i}\left(
          \Quality{\varphi_2}(\xi, j, \epsilon, \zeta) \land \bigwedge_{j \leq k \leq i}
          \Quality{\varphi_1}(\xi, k, \epsilon, \zeta) \right)
          \\
        \end{align*}

  \item For constraints on time and frame variables,
        \begin{align*}
          \Quality{x - \CTIME \sim c}(\xi, i, \epsilon, \zeta)
           & =
          \begin{cases}
            \top,\quad\text{if } \epsilon(x) - t(\xi_i) \sim c \\
            \bot,\quad\text{otherwise.}
          \end{cases}
          \\
          \Quality{f - \CFRAME \sim c}(\xi, i, \epsilon, \zeta)
           & =
          \begin{cases}
            \top,\quad\text{if } \epsilon(f) - i \sim c \\
            \bot,\quad\text{otherwise.}
          \end{cases}
          \\
        \end{align*}

  \item For operations on object variables,
        \begin{align*}
          \Quality{\{id_j = id_j\}}(\xi, i, \epsilon, \zeta)
           & =
          \begin{cases}
            \top,\quad\text{if } \zeta(id_j) = \zeta(id_k) \\
            \bot,\quad\text{otherwise.}
          \end{cases}
          \\
          \Quality{\idcls(id_j) = c}(\xi, i, \epsilon, \zeta)
           & =
          \begin{cases}
            \top,\quad\text{if } \Oc(\xi_i)(\zeta(id_j)).\textrm{class} = c \\
            \bot,\quad\text{otherwise.}
          \end{cases}
          \\
          \Quality{\idcls(id_j) = \idcls(id_k)}(\xi, i, \epsilon, \zeta)
           & =
          \begin{cases}
            \top,\quad\text{if } \Oc(\xi_i)(\zeta(id_j)).\textrm{class} \\
            \qquad\qquad = \Oc(\xi_i)(\zeta(id_k)).\textrm{class}       \\
            \bot,\quad\text{otherwise.}
          \end{cases}
          \\
          \Quality{\idprob(id_j) \sim r}(\xi, i, \epsilon, \zeta)
           & =
          \begin{cases}
            \top,\quad\text{if } \Oc(\xi_i)(\zeta(id_j)).\textrm{prob} \sim r \\
            \bot,\quad\text{otherwise.}
          \end{cases}
          \\
          \Quality{\idprob(id_j) \sim r \times \idprob(id_k)}(\xi, i, \epsilon, \zeta)
           & =
          \begin{cases}
            \top,\quad\text{if } \Oc(\xi_i)(\zeta(id_j)).\textrm{prob} \sim r \\
            \qquad\qquad\times \Oc(\xi_i)(\zeta(id_k)).\textrm{prob}          \\
            \bot,\quad\text{otherwise.}
          \end{cases}
        \end{align*}

  \item For the area, latitudinal offset, and longitudinal offset,
        \begin{align*}
          \Quality{\Area(\Tc_1) \sim r}
           & =
          \begin{cases}
            \top,\quad\text{if } \Area(\SpEval(\Tc_1, \xi, \zeta)) \sim r \\
            \bot,\quad\text{otherwise.}
          \end{cases}
          \\
          \Quality{\Lat(id_1, \CRT_1) \sim r}(\xi, i, \epsilon, \zeta)
           & =
          \begin{cases}
            \top,\quad\text{if } f_{lat}{}(id_1, \CRT_1, \xi, i, \epsilon, \zeta) \sim r \\
            \bot,\quad\text{otherwise.}
          \end{cases}
          \\
          \Quality{\Lon(id_1, \CRT_1) \sim r}(\xi, i, \epsilon, \zeta)
           & =
          \begin{cases}
            \top,\quad\text{if } f_{lon}(id_1, \CRT_1, \xi, i, \epsilon, \zeta) \sim r \\
            \bot,\quad\text{otherwise.}
          \end{cases}
        \end{align*}
        where, \(\sim \in \Set{<,>,\leq,\geq}\), and
        \begin{itemize}
          \item \(f_{lat}\) computes
                the \emph{lateral distance} of the \(\CRT\) point of an object identified by
                \(\Oc(\zeta(id_1))\) from the \emph{Longitudinal axis}; \item \(f_{lon}\) computes the
                \emph{longitudinal distance} of the \(\CRT\) point of an object identified by
                \(\Oc(\zeta(id_1))\) from the \emph{Lateral axis}; and \item \(\SpEval(\Tc, \xi,
                \zeta)\) is the compound spatial object created after set operations on bounding boxes
                (defined below).
        \end{itemize}

  \item And, finally, for the spatial existence operator,
        \begin{align*}
          \Quality{\SpExists \Tc}(\xi, i, \epsilon, \zeta)
           & =
          \begin{cases}
            \top,\quad\text{if } \SpEval(\Tc, \xi, \zeta) \not= \emptyset \\
            \bot,\quad\text{otherwise.}
          \end{cases}
        \end{align*}
        Here, the compound spatial function, \(\SpEval\) is defined as follows:
        \begin{align*}
          \SpEval(\varnothing, \xi, \zeta) & = \emptyset
          \\
          \SpEval(\Universal, \xi, \zeta)  & = \Universal                                                 \\
          \SpEval(\BBox(id), \xi,
          \zeta)                           & = \zeta(id).\textrm{bbox}                                    \\
          \SpEval(\Comp\Tc, \xi, \zeta)    & =
          \Universal \setminus \SpEval(\Tc, \xi, \zeta)                                                   \\
          \SpEval(\Tc_1 \sqcup \Tc_2, \xi,
          \zeta)                           & = \SpEval(\Tc_1, \xi, \zeta) \cup \SpEval(\Tc_2, \xi, \zeta)
        \end{align*}

\end{itemize}

\end{document}